\documentclass[runningheads]{llncs}
\usepackage{graphicx}
\usepackage{amsmath}
\usepackage{amssymb}
\usepackage{multirow}
\usepackage{booktabs}
\usepackage{hyperref}
\usepackage{color}
\usepackage{appendix}
\usepackage{cite}

\begin{document}
\begin{sloppypar}
\title{Awaker2.5-VL: Stably Scaling MLLMs with Parameter-Efficient Mixture of Experts}
\titlerunning{Awaker2.5-VL: Stably Scaling MLLMs with MoE}
%
\author{Jinqiang Long\inst{1}\thanks{They have made equal contribution and done this work during internship.} \and
Yanqi Dai\inst{1}$^\star$ \and
Guoxing Yang\inst{1} \and
Hongpeng Lin \inst{1} \and
Nanyi Fei \inst{1} \and
Yizhao Gao \inst{1} \and
Zhiwu Lu \inst{2}}
\authorrunning{Long et al.}
\institute{Metabrain AGI Lab, Shanghai, China \\
\url{https://www.metabrainagi.com}\and
Gaoling School of Artificial Intelligence, Renmin University of China
}
%
\maketitle              

\begin{abstract}
As the research of \textbf{M}ultimodal \textbf{L}arge \textbf{L}anguage \textbf{M}odels (MLLMs) becomes popular, an advancing MLLM model is typically required to handle various textual and visual tasks~(e.g., VQA, Detection, OCR, and ChartQA) simultaneously for real-world applications. However, due to the significant differences in representation and distribution among data from various tasks, simply mixing data of all tasks together leads to the well-known``multi-task conflict" issue, resulting in performance degradation across various tasks. To address this issue, we propose Awaker2.5-VL, a Mixture of Experts~(MoE) architecture suitable for MLLM, which acquires the multi-task capabilities through multiple sparsely activated experts. To speed up the training and inference of Awaker2.5-VL, each expert in our model is devised as a low-rank adaptation (LoRA) structure. Extensive experiments on multiple latest benchmarks demonstrate the effectiveness of Awaker2.5-VL. The code and model weight are released in our Project Page: \url{https://github.com/MetabrainAGI/Awaker}.

\end{abstract}
\section{Introduction}
\label{sec:intro}

With the rapid development of Large Language Model~(LLM)~\cite{bai2023qwen, llama, yi34b}, Multimodal Large Language Model~(MLLM)~\cite{blip2, llava, MiniGPT-4, internvl, minigemini}  have also become a new research hotspot in recent years. Series of MLLMs such as BLIP2~\cite{blip2}, MiniGPT-4~\cite{MiniGPT-4}, and LLaVA~\cite{llava} have demonstrated impressive performance in various vision-text tasks (e.g., image captioning, and visual question answering). Qwen-VL-Chat~\cite{qwen-vl} transfers traditional vision tasks (e.g., object detection, and OCR) to vision-text tasks, endowing the model with the ability to perform more tasks by defining specific instructions and output formats.
\vspace{0.1cm}

\noindent However, the above mentioned visual-center tasks have significant differences in image input, instructions, and output formats. For example, image captioning requires the model to perceive the entire image and generate a natural language description of its content. In contrast, object detection requires the model to locate specific objects in the image and output their exact positions in numerical coordinates. The common training strategy is to mix the training data from multiple tasks and feed it uniformly into the model for training. Since the current model architecture does not specifically address the differences among multiple tasks, this simple mixing strategy often leads to the well-known ``multi-task conflict" issue, which further results in reduced performance across all tasks.
\vspace{0.1cm}

\noindent To address the ``multi-task conflict" issue, we propose Awaker2.5-VL, a stable Mixture of Experts (MoE) architecture suitable for large multimodal models. Specifically, we set up multiple expert models to acquire the task-specific capabilities across various tasks, with a gating network automatically controlling the activation and deactivation of experts. In the meantime, we include a global expert that remains always active to ensure the versatility and generalization of the model. To speed up the training and inference of Awaker2.5-VL, each expert in our model is devised as a low-rank adaptation (LoRA) structure. Additionally, we uniformly design the routing strategy for MoE in our model. It is worth noting that during the training of Awaker2.5-VL, the base model is frozen, and only the MoE/Lora modules are trained, resulting in an extremely low training cost. Finally, we implement Awaker2.5-VL with Qwen2-VL-7B-Instruct~\cite{Qwen2VL} as the base model, and achieve state-of-the-art results on several recent benchmarks, demonstrating the effectiveness of our Awaker2.5-VL.
\vspace{0.1cm}

\noindent The main contributions of this work are summarized as follows:
\vspace{0.05cm}

\noindent \textbf{(1)} We design a stable Mixture of Experts (MoE) architecture, called Awaker2.5-VL, suitable for large multimodal models.
\vspace{0.05cm}

\noindent \textbf{(2)} We conduct an extensive exploration of MoE routing strategies and design a simple yet effective routing strategy for our proposed Awaker2.5-VL.
\vspace{0.05cm}

\noindent \textbf{(3)} We achieve state-of-the-art results on several recent benchmarks with our proposed Awaker2.5-VL.

\section{Related work}

\textbf{Multimodal Large Language Model.} In recent years, with the rapid development of large language models, large multimodal models have also become a new research hotspot, leading to many innovative research outcomes. BLIP-2~\cite{blip2} introduces the Q-Former module between the vision encoder and the LLM, which enhances the interaction between visual and language features through a query mechanism. LLaVA~\cite{llava} first generates multimodal image-text instruction data using GPT-4 and then performs instruction fine-tuning to train a large multimodal model for general-purpose visual and language understanding. The Qwen-VL~\cite{qwen-vl} series of models uniformly transform traditional visual tasks (such as object detection and OCR) into vision-language tasks, enabling them to perform a variety of different tasks including image captioning, visual question answering, object detection, and OCR.
\vspace{0.1cm}

\noindent \textbf{Mixture of Experts.} Scaling Law~\cite{scaling-law} indicates that the number of parameters in a model is typically directly related to the model's complexity and expressive power. However, simply increasing the number of parameters often leads to higher resource consumption. To reduce the model training cost, Mixture of Experts (MoE)~\cite{cai2024surveymixtureexperts} models have been introduced into large language models. MoE is a sparse, gate-controlled deep learning model primarily composed of a set of expert models and a gating model. The fundamental idea of MoE is to partition the input data into multiple regions based on task types and assign each region to one or more expert models. Each expert model can focus on processing its specific portion of the input data, thereby improving the overall performance of the model. Mistral-8x7B~\cite{mistral-8x7b} released by Mistral AI is an MoE model composed of 7 billion-parameter sub-models. It outperforms the 70 billion-parameter Llama-2 on multiple benchmarks. MOE-LLaVA~\cite{llava} proposes a sparse, multimodal large model based on MoE, which uses only about 3 billion sparsely activated parameters. Despite this, MoE-LLaVA performs comparably to LLaVA-1.5-7B on various visual understanding datasets and even surpasses LLaVA-1.5-13B on object hallucination benchmarks.

\section{Methodology}

\subsection{MoE Structure}
\begin{figure}[t!]
  \centering
  \begin{minipage}[b]{0.45\textwidth}
    \includegraphics[width=\textwidth]{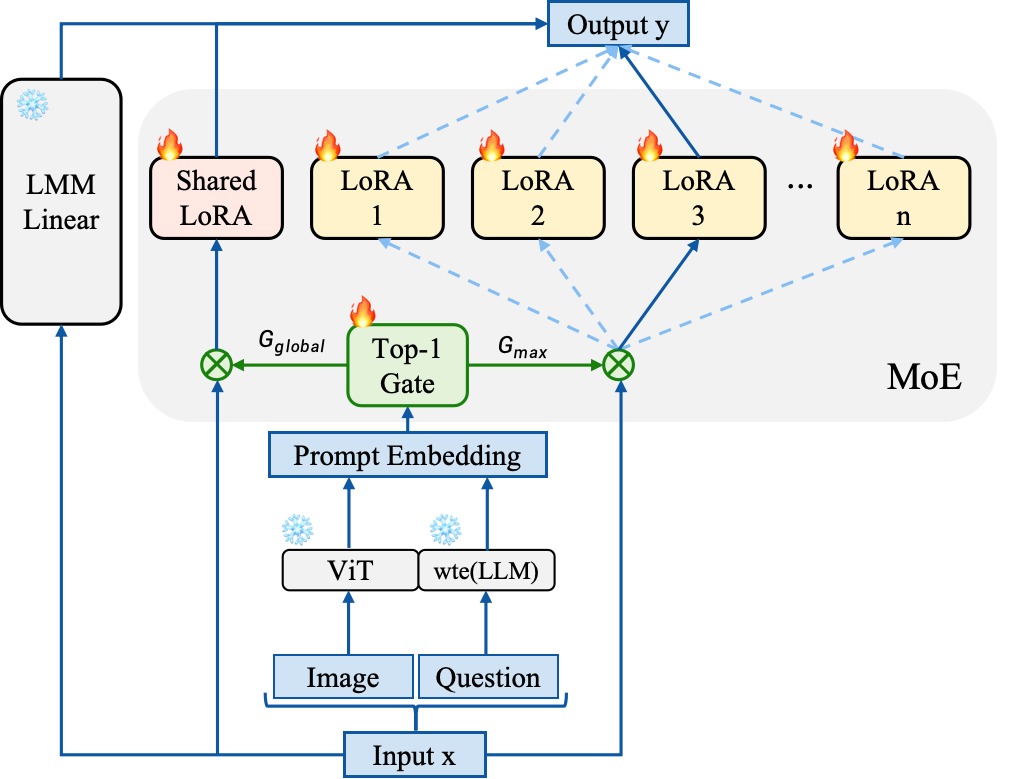}
    \caption{\bf The Standard MoE Structure in Awaker2.5-VL.}
    \label{fig:moe}
  \end{minipage}
  \hfill 
  \begin{minipage}[b]{0.45\textwidth}
    \includegraphics[width=\textwidth]{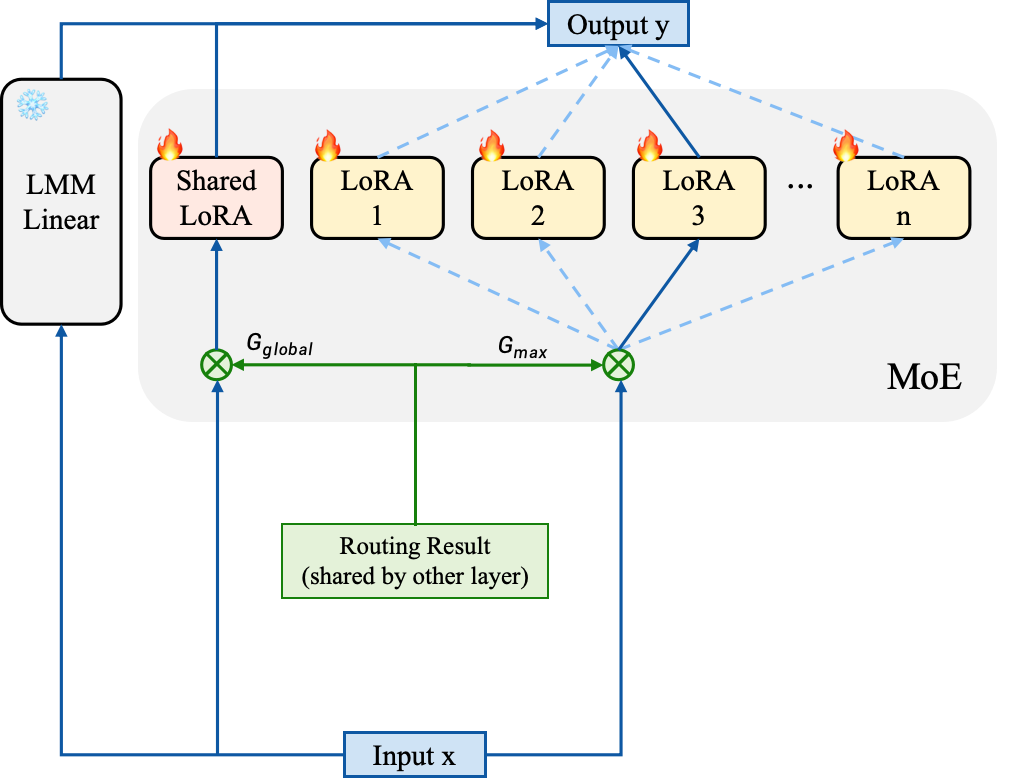}
    \caption{\bf The Simplified MoE Structure in Awaker2.5-VL.}
    \label{fig:moe-simple}
  \end{minipage}
\end{figure}

The MoE architecture of our Awaker2.5-VL~(Figure~\ref{fig:moe}), following MoCLE~\cite{mocle}, consists of a base model parameterized by $W_0$ with frozen parameters, $n$ experts and a gate layer, which can acquire the model's ability to handle various tasks. For Large Language Models, the Mixture of Experts (MoE) is typically implemented in the Feed-Forward Network layer, where each expert is an FFN layer. Unlike LLMs, each expert in Awaker2.5-VL is divesed as a LoRA structure parameterized with $W_{E}^{m}~(m \in {1,2,...,n})$. Additionally, to maintain the model's generalization capability, Awaker2.5-VL includes an always activated expert parameterized with $W_{E}^{G}$, meaning every piece of data passes through this global expert. The gate network is a simple linear layer parameterized with $W_{G} \in R^{n}$ that controls which experts are activated  when data is forwarded through the model and assigns weights to the outputs of these experts. Given an input $x$, the gating vector can be represented as follow:
\begin{equation}
    G_{\mathrm{experts}} = \mathrm{top}_k(\mathrm{softmax}(\frac{1}{\tau} (W_G x) + \epsilon))
\end{equation}

\noindent We set the number of activated experts to 1, and define $G_{\mathrm{max}}$ as the maximum value in $G_{\mathrm{experts}}$, which is used as the weight of the activated expert. The weight of the global expert can be defined as follow:
\begin{equation}
    G_{\mathrm{global}} = 1 - G_{\mathrm{max}}
\end{equation}

\noindent The final model output is determined by the combined outputs of the pre-trained model, the global expert, and the mixture of experts:
\begin{equation}
    y = \sum_{m=1}^n G_{\mathrm{experts}} W_E^m x + G_{\mathrm{global}} W_E^G x + W_0 x
\end{equation}

\noindent We also design a simplified version of the MoE architecture (as shown in Figure~\ref{fig:moe-simple}), where the gate Layer is removed. Instead, it accepts the gate results~($G_{\mathrm{global}}$ and $G_{\mathrm{experts}}$) computed in the MoE from another layer for routing. We will intersperse the use of these two types of MoE architectures throughout the model, as detailed in Chapter~\ref{subsec:imp}.


\subsection{Stable Routing Strategy}

In this work, our MoE design in Awaker2.5-VL has two main differences from traditional MoE structures in LLMs: (1) Our MoE operates at the instance-level rather than the token-level (used by traditional MoE), meaning that all tokens within each instance will activate the same expert. (2) In traditional MoE structures, the gate network of each MoE module receives the output from the previous transformer layer. However, in our practical implementation, we found that this routing strategy can lead to training instability. In this work, we thus simplify the routing strategy of MoE as shown in Figure~\ref{fig:moe-simple}.
\vspace{0.1cm}

\noindent Specifically, for a input instance, the gate layer at each transformer layer receives the output from the embedding layer of LLM as input, which keeps the same across all gate layers of Awaker2.5-VL. Moreover, to reduce the gap between training and inference, we use only the embedding of the instruction part as input to the gate network during training. The label part of the training data does not participate in the routing decision. That is, for multi-modal data, this includes the embedding of both the images and the question. For pure-text data, it is just the embedding of the question text.

\begin{figure}[t!]
\centering
\includegraphics[width=0.99\linewidth]{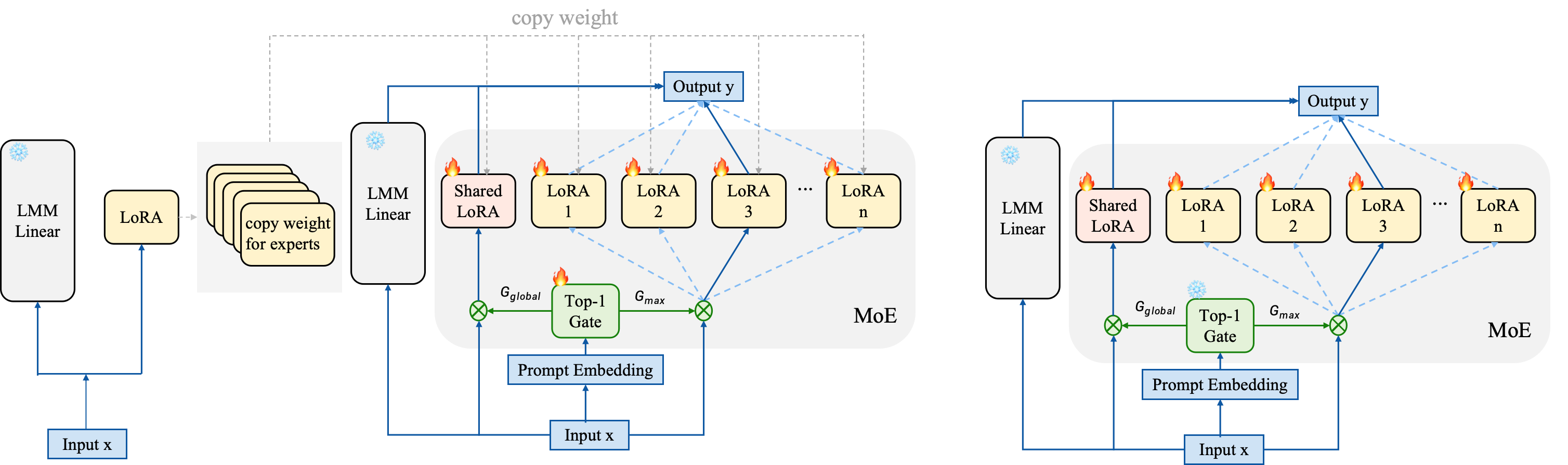}
\vspace{-0.0in}
\caption{\bf The Traing Pipeline of Awaker2.5-VL. From Left to Right: Stage \uppercase\expandafter{\romannumeral1}, Stage \uppercase\expandafter{\romannumeral2}, and Stage \uppercase\expandafter{\romannumeral3}. }
\label{fig:training}
\vspace{-0.0in}
\end{figure}

\subsection{Training Process}

We train Awaker2.5-VL through three stages as shown in Figure ~\ref{fig:training}.
\vspace{0.1cm}

\noindent \textbf{Stage \uppercase\expandafter{\romannumeral1}: Initialization Training}. In the first stage, we add a LoRA module to the base model for training. During this stage, we freeze the entire base model and only train the LoRA parameters.
\vspace{0.1cm}

\noindent \textbf{Stage \uppercase\expandafter{\romannumeral2}: MoE Training}. We further replace the LoRA module from the first stage with an MoE module in Awaker2.5-VL. Each expert in the MoE module is initialized with the LoRA parameters trained in the first stage. During this stage, we freeze the base model and only train the MoE module (including the gate layer and all the experts).
\vspace{0.1cm}

\noindent \textbf{Stage \uppercase\expandafter{\romannumeral3}: Instruction Fine-Tuning}. In the third stage, we freeze the gate layer of the MoE module and only train the experts.

\section{Experiments}

\subsection{Implementation Details}
\label{subsec:imp}

\textbf{Model Details.}
Our Awaker2.5-VL is based on the Qwen2-VL-7B-Instruct model, which is a multimodal large language model built on Qwen2. We integrate our MoE architecture into the Attention and MLP modules of Qwen2. However, in this work, we utilize different types of adapters (LoRA or MoE) based on the different layers in the model. Concretely, in the Attention module, we receptively inject a single LoRA to the \textbf{q\_proj}, \textbf{k\_proj}, and \textbf{v\_proj} layers, while adding an MoE module to the \textbf{o\_proj} layer. In the MLP module, we receptively add MoE modules to the \textbf{gate\_proj} and Simplified MoE modules to \textbf{up\_proj} and \textbf{down\_proj} layers. This means that the gate results from the \textbf{gate\_proj} layer are shared with the \textbf{up\_proj} and \textbf{down\_proj} layers.
\vspace{0.1cm}

\noindent \textbf{Hyperparameters.}
We set $n=4$ experts and 1 general expert in our MoE architecture, where each expert takes the hyperparameters $r=256, \alpha=512$. During the training process, due to cost constraints, we set the maximum image resolution to 1,103,872~(i.e., max\_pixels=1,103,872 in Qwen2-VL-7B-Instruct) and batch\_size=4. We adopt the cosine lr\_scheduler, and set the learning rating lr=1e-5 in Stage~\uppercase\expandafter{\romannumeral1}, lr=1e-5 Stage~\uppercase\expandafter{\romannumeral2}, and lr=5e-6 in Stage~\uppercase\expandafter{\romannumeral3}.

\begin{table*}[t!]
    \centering
    \caption{\bf Evaluation Results on MME-Realworld-CN Benchmark.}
    \label{tab:mme-cn}
    \vspace{-0.05in}
    \scalebox{0.86}{
    \tabcolsep4.5pt
    \begin{tabular}{lccccc}
    \toprule
     \multirow{2}{*}{Model} & \multirow{2}{*}{Parameters} & \multirow{2}{*}{Institutions} & \multicolumn{3}{c}{Chinese} \\
    \cline{4-6}
    &  & & Overall & Perception &  Reasoning  \\
    \midrule
    \textcolor{red}{\bf Awaker2.5-VL (ours)} & 10.8B  & Metabrain AGI & \textbf{62.7} & \textbf{67.71}  & \textbf{52.07}\\
    Qwen2-VL & 8B & Alibaba & 55.5 &  59.80 & 46.46\\
    InternVL-2 & 7B  & Shanghai AI Lab & 54.3 &  57.79 & 46.65 \\
    InternVL-Chat-V1.5 & 20B & Shanghai AI Lab & 47.9 & 49.90  & 43.74 \\
    Claude3.5 Sonnet & - & Anthropic & 47.0 & 48.25  & 44.31 \\
    Yi-VL-34B & 34B & 01.AI & 42.0 & 42.45  & 41.16 \\
    CogVLM2-Llama3-Chat & 8B & THU \& Zhipu AI & 39.8 &  38.57 & 42.25 \\
    GPT-4o & - &  OpenAI & 38.8 & 43.44  &  29.05 \\
    Mini-Gemini-34B-HD & 34B & CUHK & 38.5 & 38.31  & 38.75 \\
    Cambrian-1-8B & 8B &  NYU & 33.6 & 32.44  & 35.97 \\
    LLaVA-NeXT-Qwen-72B & 72B &  Bytedance & 30.6 & 30.02  & 31.67\\
    Gemini-1.5-Pro & - & Google & 28.1 & 36.10  & 11.14\\
    DeepSeek-VL & 7B & DeepSeek AI & 27.6 & 27.63  & 27.63\\
    GPT-4o-mini & - & OpenAI & 25.9 & 26.32  & 25.16 \\
    \bottomrule
    \end{tabular}}
    \vspace{-0.0in}
\end{table*}

\begin{table*}[t!]
    \centering
    \caption{\bf Evaluation Results on MME-Realworld Benchmark.}
    \label{tab:mme-eng}
    \vspace{-0.05in}
    \scalebox{0.86}{
    \tabcolsep4.5pt
    \begin{tabular}{lccccc}
    \toprule
     \multirow{2}{*}{Model} & \multirow{2}{*}{Parameters} & \multirow{2}{*}{Institutions} & \multicolumn{3}{c}{English} \\
    \cline{4-6}
    &  & & Overall & Perception &  Reasoning \\
    \midrule
    \textcolor{red}{\bf Awaker2.5-VL (ours)} & 10.8B  & Metabrain AGI & \textbf{60.8} & \textbf{63.14}  & 43.74\\
    LLaVA-OneVision & 8B  & Bytedance & 57.4 & 59.59  & 41.17\\
    Qwen2-VL & 8B & Alibaba & 56.5 &  58.96 & 40.39\\
    InternVL-2 & 7B  & Shanghai AI Lab & 53.5 &  55.82 & 38.74 \\
    Claude3.5 Sonnet & - & Anthropic & 51.6 & 52.90  & \textbf{44.12} \\
    InternVL-Chat-V1.5 & 20B & Shanghai AI Lab & 49.4 & 51.36  & 36.48 \\
     Mini-Gemini-34B-HD & 34B & CUHK & 45.9 & 48.05  & 31.73 \\
    GPT-4o & - &  OpenAI & 45.2 & 46.43  &  37.61 \\
    CogVLM2-Llama3-Chat & 8B & THU \& Zhipu AI & 44.6 &  45.84 & 37.25 \\
    Cambrian-1-8B & 8B &  NYU & 42.7 & 43.82  & 36.16 \\
    Gemini-1.5-Pro & - & Google & 38.2 & 39.63  & 29.19 \\
    GPT-4o-mini & - & OpenAI & 36.4 & 37.12  & 32.48 \\
    DeepSeek-VL & 7B & DeepSeek AI & 32.4 & 33.14  & 27.98\\
    Yi-VL-34B & 34B & 01.AI & 31.0 & 30.97  & 32.45 \\
    LLaVA-NeXT-Qwen-72B & 72B &  Bytedance & 28.7 & 29.01  & 27.86\\
    \bottomrule
    \end{tabular}}
    \vspace{-0.1in}
\end{table*}

\subsection{Training Data}

For model training, we construct a dataset of approximately 12 million pieces of data, which includes 7 million pieces of English data and 5 million pieces of Chinese data. The English data mainly comes from open-source datasets and has been filtered and selected. This primarily includes Cambrain~(2M), LLaVA-OneVision~(4M), Infinity-MM~(800K), MathV360k~(360K). The Chinese data mainly comes from our self-built dataset. In the past two years, we have carefully built a Chinese instruction dataset of about 5 million pieces of data, which includes abundant instruction data from various multimodal tasks such as Image Caption, VQA, Object Detection, and OCR.

\begin{table*}[t!]
    \centering
    \caption{\textbf{Evaluation Results on MMBench-CN Benchmark}.}
    \label{tab:mmbench-cn}
    \vspace{-0.1in}
    \scalebox{0.86}{
    \tabcolsep2.5pt
    \begin{tabular}{lccccc}
    \toprule
     \multirow{2}{*}{Model} & \multirow{2}{*}{Parameters} & \multirow{2}{*}{Institutions} & \multicolumn{3}{c}{Chinese} \\
    \cline{4-6}
    &  & & Overall & MMBench\_v1.1  &  MMBench \\
    \midrule
    Qwen2-VL-72B & 73.4B & Alibaba & \textbf{86.3}  & \textbf{85.8}  & \textbf{86.7} \\
    InternVL2-40B & 40B & Shanghai AI Lab & 85.7  & 84.9  & 86.4\\
    InternVL2-Llama-76B & 76B & Shanghai AI Lab & 85.5  & 85.5  & -\\
    Taiyi & - & Megvii & 85.2  & 85.0  & 85.4 \\
    JT-VL-Chat-V3.0 & - & China Mobile & 84.7  & 83.5  & 85.8 \\
    LLaVA-OneVision-72B & 73B & ByteDance & 84.6  & 83.9  & 85.3 \\
    Step-1.5V & - & StepFun & 84.0  & 83.5  & 84.5 \\
    Claude3.5-Sonnet-20241022 & - & Anthropic & 83.0  & 82.5  & 83.5 \\
    \textcolor{red}{\bf Awaker2.5-VL (ours)} & 10.8B & Metabrain AGI & 82.6  & 81.8  & 83.4 \\
    GPT-4o (0513, detail-low) & - & OpenAI & 82.3  & 82.5  & 82.1 \\
    LLaVA-OneVision-7B & 8B & ByteDance & 81.8  & 80.9  & 82.7 \\
    GPT-4o (0513, detail-high) & - & OpenAI & 81.8 & 81.5 & 82.1 \\
    InternVL2-26B & 26B & Shanghai AI Lab & 81.5  & 80.9  & 82.1 \\
    CongROng & - & CloudWalk & 81.2  & 80.4  & 81.9 \\
    MMAlaya2 & 26B & DataCanvas & 80.9  & 79.7  & 82.1 \\
    Ovis1.6-Gemma2-9B & 10.2B & Alibaba & 80.8  & 79.5  & 82.0 \\
    Qwen2-VL-7B & 8B & Alibaba & 80.5  & 80.3  & 80.6 \\
    LLaVA-OneVision-72B (SI)	& 73B & ByteDance & 80.0 & 81.9 & 78.0 \\
    InternVL-Chat-V1.5 & 26B & Shanghai AI Lab & 79.9 & 79.1 & 80.7 \\
    InternLM-XComposer2.5 & 8B & Shanghai AI Lab & 79.9 & 78.8 & 80.9 \\
    GPT-4o (0806, detail-high) & - & OpenAI & 79.8  & 79.2 & 80.3 \\
    GPT-4V (0409, detail-high) & - & OpenAI & 79.2  & 78.2 & 80.2 \\
    \bottomrule
    \end{tabular}}
    \vspace{-0.15in}
\end{table*}

\subsection{Main Results}

We conduct evaluation on the latest two multimodal large model benchmarks: \textbf{(1) MME-RealWorld}~\cite{mmerealworld}: this benchmark considers images from domains such as autonomous driving, remote sensing, video surveillance, newspapers, street views, and financial charts. It contains 29,429 annotations, covering 43 sub-tasks, with each task having at least 100 questions. \textbf{(2) MMBench}~\cite{mmbench}: this benchmark is a visual-language model evaluation benchmark developed by the OpenCompass research team. It enables a granular assessment of capabilities ranging from perception to cognition, covering 20 fine-grained evaluation dimensions including object detection, text recognition, action recognition, image understanding, and relational reasoning.
\vspace{0.1cm}

\noindent\textbf{Results on MME-Realworld.}
We make a comprehensive performance evaluation of our Awaker2.5-VL on the MME-Realworld benchmark and its Chinese version~(MME-Realworld-CN). Table~\ref{tab:mme-cn} and Table~\ref{tab:mme-eng} show the evaluation results in terms of perception, reasoning, and overall scores on both MME-Realworld and MME-Realworld-CN, respectively. All compared models are ranked by the average/overall scores. The results of the competitors are directly cited from \url{https://mme-realworld.github.io/home_page.html#leaderboard}. From the two tables, we have the following observations:
\vspace{0.1cm}

\noindent 1) Awaker2.5-VL ranks the first in overall score, perception score, and reasoning score on MME-RealWorld-CN, outperforming all other models. It is even the only one that takes the overall score over 60 on MME-Realworld-CN.
\vspace{0.1cm}

\noindent 2) Awaker2.5-VL still holds the top-1 position in the perception and overall scores on MME-Realworld, even though there is a slight decrease in the reasoning score when compared to the state-of-the-art.
\vspace{0.1cm}

\noindent 3) Awaker2.5-VL demonstrates exceptional performance in Chinese scenarios~(see Table~\ref{tab:mme-cn}), with an overall score improvement of 5 points compared with the base model Qwen2-VL-7B-Instruct, a 6-point increase in perception tasks, and a 3-point increase in reasoning tasks.
\vspace{0.1cm}

\begin{table*}[t!]
    \centering
    \caption{\textbf{Evaluation Results on MMBench Benchmark}.}
    \label{tab:mmbench-eng}
    \vspace{-0.1in}
    \scalebox{0.86}{
    \tabcolsep2.4pt
    \begin{tabular}{lccccc}
    \toprule
     \multirow{2}{*}{Model} & \multirow{2}{*}{Parameters} & \multirow{2}{*}{Institutions} & \multicolumn{3}{c}{English} \\
    \cline{4-6}
    &  & & Overall & MMBench\_v1.1  &  MMBench \\
    \midrule
    Qwen2-VL-72B  & 73.4B  & Alibaba  & \textbf{86.5}   & \textbf{86.1}   & \textbf{86.9} \\
    InternVL2-40B  & 40B  & Shanghai AI Lab  & 86.0   & 85.1   & 86.8 \\
    Taiyi  &  - & Megvii  & 85.7   & 84.7   & 86.7 \\
    InternVL2-Llama-76B  & 76B  & Shanghai AI Lab  & 85.5   & 85.5   & - \\
    LLaVA-OneVision-72B  & 73B  & ByteDance  & 85.4   & 85.0   & 85.8 \\
    JT-VL-Chat-V3.0  & -  & China Mobile  & 84.5   & 83.6   & 85.4 \\
    \textcolor{red}{\bf Awaker2.5-VL (ours)} & 10.8B  & Metabrain AGI  & 83.7   & 82.5   & 84.9 \\
    GPT-4o (0513, detail-high) & - & OpenAI & 83.2 & 83.0 & 83.4 \\
    GPT-4o (0513, detail-low)  &  - & OpenAI  & 83.2   & 83.1   & 83.3 \\
    Step-1.5V  & -  & StepFun  & 82.9   & 80.4   & 85.3 \\
    InternVL2-26B  & 26B  & Shanghai AI Lab  & 82.5   & 81.5   & 83.4 \\
    Ovis1.6-Gemma2-9B  & 10.2B  & Alibaba  & 82.5   & 81.5   & 83.4 \\
    RBDash-v1.2-72B  & 79B  & DLUT  & 82.5   & 81.7   & 83.2 \\
    Qwen2-VL-7B  & 8B  & Alibaba  & 82.4   & 81.8   & 83.0 \\
    LLaVA-OneVision-7B  & 8B  & ByteDance  & 82.1   & 80.9   & 83.2 \\
    GPT-4o (0806, detail-high) &  - & OpenAI  & 82.0   & 81.8   & 82.1 \\
    LLaVA-OneVision-72B (SI) & 73B & ByteDance & 81.9 & 83.3 & 80.5 \\
    Qwen-VL-Plus-0809  &  - & Alibaba  & 81.9   & 81.1   & 82.7 \\
    CongROng  &  - & CloudWalk  & 81.9   & 80.9   & 82.8 \\
    Claude3.5-Sonnet-20241022  & - & Anthropic  & 81.8   & 80.9   & 82.6 \\
    MMAlaya2  & 26B  & DataCanvas  & 81.6   & 80.6   & 82.5 \\
    InternVL-Chat-V1.5  & 26B  & Shanghai AI Lab  & 81.3 & 80.3 & 82.3 \\
    InternLM-XComposer2.5  & 8B  & Shanghai AI Lab  & 81.1 & 80.1 & 82.0 \\
    GPT-4V (0409, detail-high)  &  - & OpenAI  & 80.5   & 80.0 & 81.0 \\
    \bottomrule
    \end{tabular}}
    \vspace{-0.15in}
\end{table*}

\noindent\textbf{Results on MMBench.}
We compare Awaker2.5-VL with the latest competitors on four MMBench series benchmarks: MMBench, MMBench\_v1.1, MMBench\_CN, and MMBench\_v1.1\_CN. We separately present the results on the Chinese benchmarks~(MMBench\_CN and MMBench\_1.1\_CN) and the English benchmarks~(MMBench and MMBench\_v1.1) in Table~\ref{tab:mmbench-cn} and Table~\ref{tab:mmbench-eng}, respectively. All compared models are ranked by the average/overall scores. The results of the competitors are directly cited from \url{https://mmbench.opencompass.org.cn/leaderboard}. From the two tables, it can be observed that:
\vspace{0.1cm}

\noindent 1) Awaker2.5-VL ranks 7th on MMBench and 9th on MMBench-CN. The performance of Awaker2.5-VL on both benchmarks exceeds that of other models with similar parameter sizes.
\vspace{0.1cm}

\noindent 2) Compared to the base model Qwen2-VL-7B-Instruct, Awaker2.5-VL shows significant improvements on all four benchmarks.

\section{Conclusion and Future Work}

We release Awaker2.5-VL, a large multimodal Mixture of Experts (MoE) model. Our Awaker2.5-VL mitigates the ``multi-task conflict" issue in MLLM through the MoE architecture and has outperformed  the latest competitors on multiple benchmarks. Furthermore, we hope to enhance the capabilities of Awaker2.5-VL in the following areas in our ongoing research:
\vspace{0.1cm}

\noindent \textbf{(1)} The current prompt embeddings used for routing are derived from the embedding layers of ViT and LLM, respectively. We believe that these shallow embeddings have limited capability of representation, especially for text prompts. Therefore, in our future work, we will explore more suitable methods for representing prompts to improve routing performance.
\vspace{0.1cm}

\noindent \textbf{(2)} The MoE model in Awaker2.5-VL is currently applied only to the LLM side of the multimodal model. We plan to conduct further research on applying the MoE model to the ViT  as well.

%
%
%
{\small
\bibliographystyle{splncs04}
\bibliography{awaker}

\begin{thebibliography}{10}
\providecommand{\url}[1]{\texttt{#1}}
\providecommand{\urlprefix}{URL }
\providecommand{\doi}[1]{https://doi.org/#1}

\bibitem{yi34b}
AI, ., :, Young, A., Chen, B., Li, C., Huang, C., Zhang, G., Zhang, G., Li, H.,
  Zhu, J., Chen, J., Chang, J., Yu, K., Liu, P., Liu, Q., Yue, S., Yang, S.,
  Yang, S., Yu, T., Xie, W., Huang, W., Hu, X., Ren, X., Niu, X., Nie, P., Xu,
  Y., Liu, Y., Wang, Y., Cai, Y., Gu, Z., Liu, Z., Dai, Z.: Yi: Open foundation
  models by 01.ai (2024), \url{https://arxiv.org/abs/2403.04652}

\bibitem{bai2023qwen}
Bai, J., Bai, S., Chu, Y., Cui, Z., Dang, K., Deng, X., Fan, Y., Ge, W., Han,
  Y., Huang, F., et~al.: Qwen technical report. arXiv preprint arXiv:2309.16609
   (2023)

\bibitem{qwen-vl}
Bai, J., Bai, S., Yang, S., Wang, S., Tan, S., Wang, P., Lin, J., Zhou, C.,
  Zhou, J.: Qwen-vl: A versatile vision-language model for understanding,
  localization, text reading, and beyond (2023),
  \url{https://arxiv.org/abs/2308.12966}

\bibitem{cai2024surveymixtureexperts}
Cai, W., Jiang, J., Wang, F., Tang, J., Kim, S., Huang, J.: A survey on mixture
  of experts (2024), \url{https://arxiv.org/abs/2407.06204}

\bibitem{internvl}
Chen, Z., Wu, J., Wang, W., Su, W., Chen, G., Xing, S., Zhong, M., Zhang, Q.,
  Zhu, X., Lu, L., Li, B., Luo, P., Lu, T., Qiao, Y., Dai, J.: Internvl:
  Scaling up vision foundation models and aligning for generic
  visual-linguistic tasks (2024), \url{https://arxiv.org/abs/2312.14238}

\bibitem{mocle}
Gou, Y., Liu, Z., Chen, K., Hong, L., Xu, H., Li, A., Yeung, D.Y., Kwok, J.T.,
  Zhang, Y.: Mixture of cluster-conditional lora experts for vision-language
  instruction tuning (2024), \url{https://arxiv.org/abs/2312.12379}

\bibitem{mistral-8x7b}
Jiang, A.Q., Sablayrolles, A., Roux, A., Mensch, A., Savary, B., Bamford, C.,
  Chaplot, D.S., de~las Casas, D., Hanna, E.B., Bressand, F., Lengyel, G.,
  Bour, G., Lample, G., Lavaud, L.R., Saulnier, L., Lachaux, M.A., Stock, P.,
  Subramanian, S., Yang, S., Antoniak, S., Scao, T.L., Gervet, T., Lavril, T.,
  Wang, T., Lacroix, T., Sayed, W.E.: Mixtral of experts (2024),
  \url{https://arxiv.org/abs/2401.04088}

\bibitem{scaling-law}
Kaplan, J., McCandlish, S., Henighan, T., Brown, T.B., Chess, B., Child, R.,
  Gray, S., Radford, A., Wu, J., Amodei, D.: Scaling laws for neural language
  models (2020), \url{https://arxiv.org/abs/2001.08361}

\bibitem{blip2}
Li, J., Li, D., Savarese, S., Hoi, S.: Blip-2: Bootstrapping language-image
  pre-training with frozen image encoders and large language models (2023),
  \url{https://arxiv.org/abs/2301.12597}

\bibitem{minigemini}
Li, Y., Zhang, Y., Wang, C., Zhong, Z., Chen, Y., Chu, R., Liu, S., Jia, J.:
  Mini-gemini: Mining the potential of multi-modality vision language models
  (2024), \url{https://arxiv.org/abs/2403.18814}

\bibitem{llava}
Liu, H., Li, C., Wu, Q., Lee, Y.J.: Visual instruction tuning (2023),
  \url{https://arxiv.org/abs/2304.08485}

\bibitem{mmbench}
Liu, Y., Duan, H., Zhang, Y., Li, B., Zhang, S., Zhao, W., Yuan, Y., Wang, J.,
  He, C., Liu, Z., Chen, K., Lin, D.: Mmbench: Is your multi-modal model an
  all-around player? (2024), \url{https://arxiv.org/abs/2307.06281}

\bibitem{llama}
Touvron, H., Martin, L., Stone, K., Albert, P., Almahairi, A., Babaei, Y.,
  Bashlykov, N., Batra, S., Bhargava, P., Bhosale, S., et~al.: Llama 2: Open
  foundation and fine-tuned chat models. arXiv preprint arXiv:2307.09288
  (2023)

\bibitem{Qwen2VL}
Wang, P., Bai, S., Tan, S., Wang, S., Fan, Z., Bai, J., Chen, K., Liu, X.,
  Wang, J., Ge, W., Fan, Y., Dang, K., Du, M., Ren, X., Men, R., Liu, D., Zhou,
  C., Zhou, J., Lin, J.: Qwen2-vl: Enhancing vision-language model's perception
  of the world at any resolution. arXiv preprint arXiv:2409.12191  (2024)

\bibitem{mmerealworld}
Zhang, Y.F., Zhang, H., Tian, H., Fu, C., Zhang, S., Wu, J., Li, F., Wang, K.,
  Wen, Q., Zhang, Z., Wang, L., Jin, R., Tan, T.: Mme-realworld: Could your
  multimodal llm challenge high-resolution real-world scenarios that are
  difficult for humans? (2024), \url{https://arxiv.org/abs/2408.13257}

\bibitem{MiniGPT-4}
Zhu, D., Chen, J., Shen, X., Li, X., Elhoseiny, M.: Minigpt-4: Enhancing
  vision-language understanding with advanced large language models (2023),
  \url{https://arxiv.org/abs/2304.10592}

\end{thebibliography}
}
%




\end{sloppypar}

\end{document}